\documentclass[sigconf,nonacm]{acmart}

\usepackage{microtype}
\usepackage{graphicx}
\usepackage{subcaption}
\usepackage{booktabs}
\usepackage{multirow}
\usepackage{placeins}

\usepackage{mathtools}

\usepackage[capitalize,noabbrev]{cleveref}

\AtEndPreamble{%
  \theoremstyle{acmdefinition}

}

\usepackage[textsize=tiny]{todonotes}

\AtBeginDocument{%
  }

\begin{document}

\title[Token-Efficient Multimodal Reasoning via IPPg]{Token-Efficient Multimodal Reasoning via Image Prompt Packaging}

\author{Joong Ho Choi}
\email{joongho.choi@bny.com}
\affiliation{\institution{BNY}\city{Pittsburgh}\country{US}}

\author{Jiayang Zhao}
\email{jiayang.zhao@bny.com}
\affiliation{\institution{BNY}\city{Pittsburgh}\country{US}}

\author{Avani Appalla}
\email{avani.appalla@bny.com}
\affiliation{\institution{BNY}\city{Pittsburgh}\country{US}}

\author{Himansh Mukesh}
\email{himanshmukesh.mulchandani@bny.com}
\affiliation{\institution{BNY}\city{Pittsburgh}\country{US}}

\author{Dhwanil Vasani}
\email{dhwanil.vasani@bny.com}
\affiliation{\institution{BNY}\city{Pittsburgh}\country{US}}

\author{Boyi Qian}
\email{boyi.qian@bny.com}
\affiliation{\institution{BNY}\city{Pittsburgh}\country{US}}

\keywords{Machine Learning, Multimodal Reasoning, Visual Prompting, Token Efficiency}

\begin{abstract}

Deploying large multimodal language models at scale is constrained by token-based inference costs, yet the cost-performance behavior of visual prompting strategies remains poorly characterized. We introduce Image Prompt Packaging (IPPg), a prompting paradigm that embeds structured text directly into images to reduce text token overhead, and benchmark it across five datasets, three frontier models (GPT-4.1, GPT-4o, Claude 3.5 Sonnet), and two task families (VQA and code generation). We derive a cost formulation decomposing savings by token type and show IPPg achieves 35.8--91.0\% inference cost reductions. Despite token compression of up to 96\%, accuracy remains competitive in many settings, though outcomes are highly model- and task-dependent: GPT-4.1 achieves simultaneous accuracy and cost gains on CoSQL, while Claude 3.5 incurs cost increases on several VQA benchmarks. Systematic error analysis yields a failure-mode taxonomy: spatial reasoning, non-English inputs, and character-sensitive operations are most vulnerable, while schema-structured tasks benefit most. A 125-configuration rendering ablation reveals accuracy shifts of 10--30 percentage points, establishing visual encoding choices as a first-class variable in multimodal system design.

\end{abstract}

\maketitle

\section{Introduction}

Multimodal fusion is central to modern AI systems, yet the practical economics of multimodal inference remain underexplored. Token-based billing dominates the cost structure of commercial LLM APIs, and structured inputs---verbose schema descriptions, long financial prompts, dense receipts---can consume thousands of text tokens per query. As multimodal workloads scale, the question of how to reduce this overhead without sacrificing task accuracy becomes a first-order deployment concern. Despite the commercial significance of this problem, there exists no systematic benchmark characterizing the cost-performance behavior of visual prompting strategies across diverse domains, model families, and task types, including healthcare VQA, financial reasoning, document extraction, Python code generation, and multi-agent text-to-SQL.

To address this, we propose Image Prompt Packaging (IPPg), a cost-aware prompting framework that embeds structured text directly into images and routes inputs through the visual tokenization channel, eliminating text token costs for the injected content. IPPg is architecture-agnostic and requires no model modifications, making it straightforward to evaluate across commercial APIs. We benchmark IPPg across five datasets and three frontier models using two controlled experimental paradigms---(i) text+image vs. text-in-image and (ii) text vs. text-as-image---designed to isolate how visual encoding affects semantic fidelity, tokenization efficiency, and downstream accuracy. Our key contributions are

\begin{itemize}
    \item We introduce IPPg and benchmark it across five datasets, three frontier models, and two task families---VQA and code generation---including multi-agent and two-call repair settings.
    \item We derive a cost formulation that decomposes inference cost into image- and text-token components under provider-specific tokenization schemes (tile-based for OpenAI, pixel-linear for Claude), and report performance-per-dollar and cost-per-correct-answer across all configurations, metrics that better reflect commercial deployment constraints than raw token counts.
    \item Through error analysis across all five datasets, we construct a failure-mode taxonomy for visual text encoding, identifying spatial reasoning, non-English inputs, character-sensitive string manipulation, order-sensitive arithmetic, and multi-table relational reasoning as consistent bottlenecks, findings that generalize across model families and carry implications for MLLM design beyond IPPg.
    \item We conduct the first systematic rendering ablation for text-as-image prompting, evaluating 125 font $\times$ color $\times$ size combinations across three VQA datasets, demonstrating accuracy shifts of 10--30 percentage points and identifying empirically grounded configuration guidelines and model-specific accuracy-cost Pareto frontiers for practitioners.
\end{itemize}

\section{Related Work}

In multimodal learning, information across modalities is typically combined using Early Fusion or Late Fusion frameworks. Early Fusion projects modality-specific inputs into a shared embedding space, allowing a single transformer to process mixed sequences of image and text tokens without separate modality-specific encoders or decoders \cite{anil2023gemini, betker2023improving}. This approach supports unified cross-modal reasoning but often incurs high representational complexity. In contrast, Late Fusion trains independent models per modality and aggregates their predictions, reducing per-model complexity while potentially failing to capture fine-grained inter-modal interactions \cite{mmlf2024multimodal}.

However, both Early and Late Fusion approaches are susceptible to modality bias, where models over-rely on dominant modalities while under-utilizing others, leading to degraded performance when certain inputs are missing or unreliable. This bias can arise from asymmetric modality backbones, dataset imbalances, training objectives, and differing convergence dynamics across modalities \cite{zheng2025mllms}. Recent work formalizes this phenomenon through metrics such as the Modality Dominance Index (MDI) and Attention Efficiency Index (AEI), revealing a strong tendency for multimodal LLMs to prioritize textual inputs during inference at the expense of visual or auditory information \cite{wu2025when}. Some Modalities Are More Equal Than Others shows that MLLMs often fail to prioritize the queried modality under audio--visual or text--visual misalignment, frequently defaulting to textual cues even when they are misleading. These findings suggest that modality dominance is not purely semantic, but is strongly shaped by tokenizer efficiency and channel-specific inductive biases \cite{chen2025modalitiesequalothersdecoding}. IPPg operationalizes this insight as an inference-time intervention: by deliberately routing textual content through the visual channel, it stress-tests modality dominance empirically across diverse tasks, yielding the first large-scale behavioral map of when visual encoding helps or hurts.

To mitigate modality bias, prior work has proposed training-time interventions. For example, modality collapse in multimodal VAEs has been attributed to conflicting gradients across modalities, motivating gradient-conflict detection and resolution techniques adapted from multi-task learning \cite{javaloy2022mitigating}. Other approaches introduce explicit basis reallocation or cross-modal knowledge distillation to disentangle representations and alleviate rank bottlenecks in multimodal encoders \cite{chaudhuri2025closer}.

Encoding fidelity---the extent to which semantic content is preserved under cross-modal conversion---remains under-explored. In particular, embedding textual prompts directly into images and relying solely on the vision channel, rather than jointly using text and image modalities, raises open questions about semantic degradation and its downstream impact on LLM reasoning. Recently, two noteworthy papers have explored multi-modality in related directions. DeepSeek-OCR uses a $16\times$ convolutional compressor and DeepSeek-3B-MoE decoder that reconstructs text, HTML, and figure annotations with minimal loss, enabling language models to attend to images of text more cost-efficiently \cite{Wei2025DeepSeekOCRCO}. However, this is an architectural contribution requiring model modification; we are interested in a pure prompt-time strategy compatible with any off-the-shelf MLLM. Another paper shows that textual inputs can be compressed by feeding them as images to reduce token usage while preserving performance on long-context retrieval and document summarization benchmarks \cite{Li2025TextOP}. However, their evaluation is restricted to text-only-to-image conversion on two benchmarks with two models, neither of which involves genuine multimodal inputs. They also report only token-level compression without accounting for commercial cost structures where input and output tokens are priced asymmetrically. Our benchmark extends this in scope (five datasets, three models, multimodal tasks), cost modeling (provider-specific tokenization, performance-per-dollar), and rendering analysis (125 configurations vs. resolution only).

Token-based billing and throughput constraints dominate the cost structure of commercial LLMs, yet are largely overlooked in existing multimodal fusion research, limiting practical scalability. Because LLM APIs bill by token counts, languages that need more tokens to express the same content end up costing users more. \cite{ahia2023languages} illustrates that heavily segmented, non-Latin scripts, especially mid-resourced Indic languages, can be nearly five times more expensive than English, while Indo-European, Latin-script languages remain the cheapest.

Recent work on typographic attacks and visual text distraction has shown that embedded text can, in some cases, mislead or override visual reasoning in vision-language models \cite{cheng2024unveilingtypographicdeceptionsinsights}. Beyond issues of relevance and distractibility, recent studies also reveal fundamental limitations of current vision-language models in fine-grained text understanding. For instance, Li et al.\ find that even strong VLMs struggle with tasks such as font recognition, indicating that semantic and stylistic understanding of text remains a challenging open problem despite advances in broader multimodal reasoning \cite{li2025texturesemanticsvisionlanguagemodels}. While many earlier evaluation protocols implicitly encouraged models to disregard embedded text, Waseda et al.\ show that realistic image understanding often requires joint reasoning over both visual objects and textual cues, highlighting the need for context-aware text handling in multimodal systems \cite{waseda2025readignoreunifiedbenchmark}. To address this, Cheng et al.\ introduce a large-scale typographic dataset to systematically study the distractibility of LVLMs by text in images, and demonstrate that more informative text prompts can partially mitigate the resulting performance degradation \cite{cheng2024unveilingtypographicdeceptionsinsights}.

Our work provides a comprehensive examination of encoding fidelity by embedding textual prompts of simple, clear Arial font directly into images and routing them through the vision channel across a broad range of domains and tasks. In doing so, we explicitly connect semantic preservation to real-world commercial constraints, including token-based pricing and inference cost.

\section{Design}

In our ablation study, we selected GPT-4.1 \cite{openai_chatgpt4.1}, GPT-4o \cite{openai_chatgpt4o_omni}, and Claude 3.5 Sonnet \cite{claude_sonnet_3.5} to ensure broad, vendor-agnostic coverage across the leading edge of commercial MLLMs. We use the Python Image Library (PIL) with a default DPI of 72 to inject the text prompt into the image using the Arial font. For different image sizes, we automatically adjust the injected prompt text to follow the original width of the image. To ensure fairness across experiments, we employ a vanilla LLM configuration with no tool calls or external skills enabled. For tasks that require a specific output format, we keep the system prompt to state the output format for alignment, while we inject the user prompt into the image. We then provide the model both the system prompt and the rendered image.

\subsection{Cost Profiling}

\begin{table}[h]
\centering
\caption{Per-token pricing by model (\$/token). All models satisfy $p_{\mathrm{img}} = p_i$.}
\label{tab:pricing}
\setlength{\tabcolsep}{6pt}
\begin{tabular}{lccc}
\toprule
\textbf{Model} & $p_i$ & $p_o$ & $p_o/p_i$ \\
\midrule
GPT-4o~\cite{openai_pricing} & $2.50 \times 10^{-6}$ & $1.00 \times 10^{-5}$ & 4 \\
GPT-4.1~\cite{openai_pricing} & $2.00 \times 10^{-6}$ & $8.00 \times 10^{-6}$ & 4 \\
Claude 3.5 Sonnet~\cite{anthropic_claude_vision} & $3.00 \times 10^{-6}$ & $1.50 \times 10^{-5}$ & 5 \\
\bottomrule
\end{tabular}
\end{table}

Let $n_i$ and $n_o$ denote the numbers of input and output text tokens, and let $n_{\mathrm{img}}^{\mathrm{base}}$ and $n_{\mathrm{img}}^{\mathrm{IPPg}}$ denote the numbers of image tokens for the baseline image and the IPPg image (with embedded text), respectively. Let $p_i$, $p_o$, and $p_{\mathrm{img}}$ be the per-token prices for input text, output text, and image tokens. Table~\ref{tab:pricing} summarizes the per-token pricing for each model (as of January 2026); across providers, image tokens are priced identically to input text tokens, i.e., $p_{\mathrm{img}} = p_i$.

To isolate the effect of IPPg on inference cost, we condition on a fixed task prompt and decoding configuration and assume that both methods incur the same text token counts. Specifically, $n_i^{\mathrm{baseline}} = n_i^{\mathrm{IPPg}}$ and $n_o^{\mathrm{baseline}} = n_o^{\mathrm{IPPg}}$, so any cost difference arises solely from the image token counts ($n_{\mathrm{img}}^{\mathrm{base}}$ versus $n_{\mathrm{img}}^{\mathrm{IPPg}}$).

{\setlength{\abovedisplayskip}{4pt}
 \setlength{\belowdisplayskip}{4pt}
 \setlength{\abovedisplayshortskip}{2pt}
 \setlength{\belowdisplayshortskip}{2pt}

The baseline approach (text + image) incurs cost:
\begin{equation}
\mathrm{Cost}_{\mathrm{baseline}} = p_i \, n_i + p_{\mathrm{img}} \, n_{\mathrm{img}}^{\mathrm{base}} + p_o \, n_o.
\end{equation}
Our IPPg approach (text embedded in image) incurs cost:
\begin{equation}
\mathrm{Cost}_{\mathrm{IPPg}} = p_{\mathrm{img}} \, n_{\mathrm{img}}^{\mathrm{IPPg}} + p_o \, n_o.
\end{equation}

\label{thm:cost_advantage}
Let $\Delta n_{\mathrm{img}} = n_{\mathrm{img}}^{\mathrm{IPPg}} - n_{\mathrm{img}}^{\mathrm{base}}$ denote the additional image tokens incurred by embedding text into the image. Under the assumption $p_{\mathrm{img}} = p_i$, IPPg is strictly cheaper than the baseline if and only if
\begin{equation}
\Delta n_{\mathrm{img}} < n_i.
\end{equation}
That is, embedding text into an image saves cost whenever the additional image tokens are fewer than the text tokens eliminated.

IPPg is cheaper when $\mathrm{Cost}_{\mathrm{IPPg}} < \mathrm{Cost}_{\mathrm{baseline}}$:
\[
p_{\mathrm{img}} \, n_{\mathrm{img}}^{\mathrm{IPPg}} + p_o \, n_o < p_i \, n_i + p_{\mathrm{img}} \, n_{\mathrm{img}}^{\mathrm{base}} + p_o \, n_o.
\]
The output cost $p_o \, n_o$ cancels from both sides:
\[
p_{\mathrm{img}} \, n_{\mathrm{img}}^{\mathrm{IPPg}} < p_i \, n_i + p_{\mathrm{img}} \, n_{\mathrm{img}}^{\mathrm{base}}.
\]
Since $p_{\mathrm{img}} = p_i$, we can factor out $p_i$:
\[
n_{\mathrm{img}}^{\mathrm{IPPg}} < n_i + n_{\mathrm{img}}^{\mathrm{base}}.
\]
Rearranging gives $n_{\mathrm{img}}^{\mathrm{IPPg}} - n_{\mathrm{img}}^{\mathrm{base}} < n_i$; i.e., $\Delta n_{\mathrm{img}} < n_i$.

[Cost Savings]
When $\Delta n_{\mathrm{img}} < n_i$, the per-query cost savings are:
\[
\mathrm{Savings} = p_i \, (n_i - \Delta n_{\mathrm{img}}).
\]

[Modal Token Equivalence]
Since $p_{\mathrm{img}} = p_i$ across all tested models:
\[
1 \text{ image token} \equiv 1 \text{ input text token}
\]
in terms of marginal cost. Output tokens are $4\times$ (GPT) or $5\times$ (Claude) more expensive than input tokens.

It is worth noting that for OpenAI models, image tokens are computed via a tile-based formula:
\begin{equation}
n_{\mathrm{img}} = T_{\mathrm{base}} + T_{\mathrm{tile}} \cdot \lceil W/512 \rceil \cdot \lceil H/512 \rceil,
\end{equation}
where $T_{\mathrm{base}} = 85$ and $T_{\mathrm{tile}} = 170$~\cite{openai_image_pricing}; for Claude, image tokens scale linearly with pixel area~\cite{anthropic_claude_vision}:

\begin{equation}
n_{\mathrm{img}} = \frac{W \cdot H}{750}.
\end{equation}

In practice, embedding text into an image often adds minimal pixels (e.g., a text banner), which may not increase the tile count when the rendered image remains within the same post-resize tile grid for OpenAI models. When $\Delta n_{\mathrm{img}} \approx 0$, the full text token cost $p_i \, n_i$ is saved.
}

\subsection{IPPg Pipeline}

To initiate IPPg, we take an image, create minimal necessary whitespace above the image, and inject the text prompt into the whitespace.

With this approach, we not only remove the consideration of text token billing but also minimize the increase in image tokens. Then, we compare the performance and cost to the baseline approach, which involves sending both prompt and image and thus incurring both text and image token costs.

\section{Downstream Tasks}

We evaluate IPPg across multiple downstream tasks that span vision-language reasoning, document understanding, and long-context code generation. This diversity helps identify when IPPg is most effective and characterize how its benefits depend on modality balance and model architecture. Across all tasks, we report task-specific accuracy metrics alongside token-based billing cost, enabling a direct cost-performance comparison with standard text-based prompting.

\subsection{FAMMA: Financial Multimodal Question Answering}

FAMMA is a large-scale multimodal benchmark for conversational financial question answering that requires multi-step numerical and logical reasoning in domain-specific terminology. Each query includes text along with up to seven visual inputs (e.g., tables, charts, and document screenshots), mirroring real-world analyst workflows.

FAMMA is well suited for evaluating IPPg because prompts are verbose while visual inputs are already central. We compare standard text+image prompting with text-in-image, measuring accuracy and inference cost across models, and report results by question type (multiple-choice vs.\ open-ended) and language (English vs.\ non-English) to assess sensitivity to prompt structure and linguistic variation.

\subsection{PathVQA: Medical Visual Question Answering}

PathVQA evaluates question answering over pathology images containing complex biomedical structures and no embedded text. This makes it well suited for testing whether embedding textual prompts into images interferes with visual reasoning. We evaluate IPPg by rendering the question prompt directly into the image and comparing it against standard text prompting. Performance is measured by answer accuracy, while efficiency is assessed using image cost metrics.

\subsection{SROIE: Receipt Understanding and Information Extraction}

SROIE benchmarks information extraction from scanned receipts, including vendor names, dates, and monetary values. The dataset exhibits high variability in layout, text density, and image quality, reflecting real-world document understanding challenges. We use SROIE to assess whether image-embedded prompts affect field-level extraction accuracy under IPPg. Evaluation focuses on numerical fields (cash and change), reporting exact-match accuracy, parsability, and cost per correct extraction.

\subsection{HumanEval: Python Code Generation}

HumanEval is a standard benchmark for Python code generation, consisting of 164 programming tasks with correctness evaluated via unit tests. Although the task is purely text-based, it enables analysis of IPPg without native visual input. We compare standard text prompting to rendering the full problem description as an image. Performance is measured using pass@1 accuracy under single-call and two-call repair settings, along with output token length and total inference cost to assess conciseness and efficiency.

\subsection{CoSQL with MAC-SQL: Conversational Text-to-SQL}

CoSQL is a large-scale conversational text-to-SQL dataset spanning multiple domains and complex database schema; it has 3007 questions. We evaluate using the MAC-SQL framework, which decomposes SQL generation into three stages: SELECTOR (table selection), DECOMPOSER (query generation), and REFINER (error correction).

This task is especially well suited for IPPg because schema descriptions are long and token-intensive. We replace text-based schema inputs with image-rendered schemas in the SELECTOR and DECOMPOSER stages, while keeping the REFINER stage text-based. Performance is measured using exact-match execution accuracy, while efficiency is evaluated via per-query cost and agent-level token usage. This setting allows us to isolate where text-as-image encoding yields the largest cost savings.

\section{Result}

\subsection{IPPg Application for VQA}

\subsubsection{FAMMA}

\begin{table*}[h]
\centering
\caption{FAMMA results by question type and language. Cost savings computed from average per-query costs.}
\label{tab:famma_breakdown}
\setlength{\tabcolsep}{4pt}
\small
\begin{tabular}{llcccccc}
\toprule
\textbf{Model} & \textbf{Slice} &
\textbf{Base Acc} & \textbf{IPPg Acc} &
\textbf{$\Delta$Acc} &
\textbf{Text \$} & \textbf{Img \$} &
\textbf{Save} \\
\midrule
\multirow{6}{*}{GPT-4.1}
 & Overall & 39.21\% & 30.24\% & -8.97\% & 0.001644 & 0.001247 & 24.1\% \\
 & MCQ & 39.72\% & 34.37\% & -5.35\% & 0.001172 & 0.001015 & 13.4\% \\
 & Open-Ended & 38.61\% & 25.41\% & -13.20\% & 0.002197 & 0.001519 & 30.9\% \\
 & English & 38.27\% & 34.36\% & -3.91\% & 0.001609 & 0.001227 & 23.7\% \\
 & Non-English & 41.86\% & 18.60\% & -23.26\% & 0.001742 & 0.001305 & 25.1\% \\
 & Eng.+MCQ & 38.35\% & 37.63\% & -0.72\% & 0.001160 & 0.000991 & 14.6\% \\
\midrule
\multirow{6}{*}{GPT-4o}
 & Overall & 34.50\% & 24.62\% & -9.88\% & 0.001776 & 0.001398 & 21.3\% \\
 & MCQ & 41.41\% & 30.99\% & -10.42\% & 0.001489 & 0.001249 & 16.1\% \\
 & Open-Ended & 26.40\% & 17.16\% & -9.24\% & 0.002113 & 0.001572 & 25.6\% \\
 & English & 35.32\% & 29.36\% & -5.95\% & 0.001759 & 0.001373 & 21.9\% \\
 & Non-English & 32.16\% & 11.11\% & -21.05\% & 0.001827 & 0.001470 & 19.5\% \\
 & Eng.+MCQ & 41.07\% & 36.43\% & -4.64\% & 0.001483 & 0.001230 & 17.1\% \\
\midrule
\multirow{6}{*}{Claude 3.5}
 & Overall & 18.26\% & 16.59\% & -1.67\% & 0.002567 & 0.003276 & -27.6\% \\
 & MCQ & 25.35\% & 21.97\% & -3.38 \% & 0.001974 & 0.003156 & -59.9\% \\
 & Open-Ended & 9.93\% & 10.26\% & +0.33\% & 0.003263 & 0.003417 & -4.7\% \\
 & English & 16.67\% & 15.00\% & -1.67\% & 0.002545 & 0.003315 & -30.3\% \\
 & Non-English & 35.09\% & 33.33\% & -1.75\% & 0.002793 & 0.002859 & -2.4\% \\
 & Eng.+MCQ & 23.64 \% & 19.70\% & -3.94\% & 0.001979 & 0.003174 & -60.4\% \\
\bottomrule
\end{tabular}
\end{table*}

We evaluated IPPg across \textbf{658} questions (\textbf{355 multiple-choice}, \textbf{303 open-ended}) in English and non-English, comparing the text+image baseline against IPPg on three models: GPT-4.1, GPT-4o, and Claude Sonnet 3.5. Overall, GPT-4.1 and GPT-4o achieved substantial cost savings (21--24\%) at the expense of a 9--10 percentage-point drop in aggregate accuracy, while Claude Sonnet 3.5 saw costs rise by 28\% despite only a 1.7 percentage-point accuracy change, as shown in Table~\ref{tab:famma_breakdown}. Results varied significantly by question type and language: multiple-choice questions and English inputs fared best, whereas open-ended or non-English prompts incurred larger accuracy penalties.

Across all question types and languages, GPT-4.1's IPPg approach reduced costs by \textbf{24.1\%}, but accuracy fell from 39.21\% to 30.24\% (-8.97\%). When broken down further, multiple-choice prompts lost only 5.35 percentage points while saving 13.4\% in cost; open-ended questions lost 13.2 percentage points for 30.9\% savings. English-only inputs yielded a modest 3.91 percentage-point accuracy drop with 23.7\% cost savings. Notably, the optimal configuration---English multiple-choice---achieved a mere 0.72 percentage-point accuracy decline (38.35\% $\to$ 37.63\%) while cutting costs by 14.6\%, making GPT-4.1 a strong deployment candidate.

GPT-4o showed similar trends: overall accuracy dropped from 34.50\% to 24.62\% (-9.88\%) with \textbf{21.3\%} cost savings. Under multiple-choice scenarios, it sacrificed 10.42 percentage points of accuracy to save 16.1\% in cost, and on open-ended tasks lost 9.24 percentage points for 25.6\% savings. English questions fared better (-5.95\% accuracy, 21.9\% savings), and in the English + multiple-choice setting GPT-4o saw a 4.64\% accuracy drop alongside a 17.1\% cost reduction. Despite attractive savings, the steeper accuracy hit relative to GPT-4.1 suggests more cautious deployment.

In contrast, Claude Sonnet 3.5 fails to benefit from IPPg: overall accuracy dipped just 1.67\% (18.26\% $\to$ 16.59\%), but costs \emph{increased} by 27.6\%. On multiple-choice, it lost 3.38\% accuracy while costs jumped 59.9\%; on open-ended tasks accuracy actually rose 0.33\%, but costs still rose 4.7\%. English prompts saw a 1.67\% accuracy drop with a 30.3\% cost increase. Thus, Claude 3.5 is not a viable candidate for this optimization.

Our error analysis reveals striking architectural patterns: GPT-4.1 achieves superior cross-model agreement between GPT variants, coupled with identical token efficiency ratios, indicating shared architectural characteristics in their approach to financial analysis. Claude's distinct behavior, marked by significantly longer responses and higher token usage, suggests that it attempts to decompose complex financial problems through step-by-step reasoning, whereas GPT models employ a more direct pattern-matching approach. This fundamental difference in problem-solving strategy becomes particularly apparent in failure cases, where Claude's verbose outputs show multiple attempted solution paths while GPT models maintain concise, single-path responses, revealing contrasting approaches to handling uncertainty in financial reasoning tasks.

\textbf{Overall, as long as the questions are not particularly open-ended or in non-English languages, IPPg may be preferable for cost-sensitive situations with GPT-4o and GPT-4.1, given a reasonable performance-cost tradeoff, but not with Claude 3.5.}

\subsubsection{PathVQA}

\begin{table*}[h]
\centering
\caption{Summary of IPPg across datasets. Accuracy and average inference cost are reported for text-based and image-based prompting. Cost savings are relative to text-only baselines.}
\label{tab:overall_results}
\setlength{\tabcolsep}{3pt}
\small
\begin{tabular}{llcccccc}
\toprule
\textbf{Dataset} & \textbf{Model} &
\textbf{Base Acc} & \textbf{IPPg Acc} &
\textbf{$\Delta$Acc} &
\textbf{Text \$} & \textbf{Img \$} &
\textbf{Save} \\
\midrule
\multirow{3}{*}{PathVQA} & GPT-4.1 & 42.70\% & 42.40\% & -0.30\% & 0.00290 & 0.00127 & 56.0\% \\
 & GPT-4o & 35.65\% & 36.71\% & 1.06\% & 0.001195 & 0.001128 & 5.6\% \\
 & Claude 3.5 & 39.80\% & 34.90\% & -4.90\% & 0.00140 & 0.001913 & -35.7\% \\
\midrule
\multirow{3}{*}{SROIE} & GPT-4.1 & 92.70\% & 91.00\% & -1.70\% & 0.002303 & 0.002273 & 1.2\% \\
 & GPT-4o & 87.60\% & 89.20\% & +1.60\% & 0.002865 & 0.002829 & 1.3\% \\
 & Claude 3.5 & 88.90\% & 92.30\% & +3.40\% & 0.003997 & 0.003930 & 1.5\% \\
\midrule
\multirow{3}{*}{HumanEval} & GPT-4.1 & 95.70\% & 85.40\% & -10.3\% & 0.0026 & 0.0023 & 11.50\% \\
 & GPT-4o & 91.50\% & 82.3\% & -9.20\% & 0.0039 & 0.0038 & 2.6\% \\
 & Claude 3.5 & 94.0\% & 53.0\% & -41.0\% & 0.0051 & 0.0036 & 29.4\% \\
\midrule
\multirow{3}{*}{CoSQL} & GPT-4.1 & 53.75 \% & 55.13\% & +1.38\% & 0.0740 & 0.0067 & 91.0\% \\
 & GPT-4o & 58.67\% & 54.67\% & -4.00 \% & 0.0084 & 0.0052 & 37.4\% \\
 & Claude 3.5 & 50.00 \% & 44.62 \% & -5.38 \% & 0.0117 & 0.0075 & 35.8\% \\
\bottomrule
\end{tabular}
\end{table*}

On 1,000 randomly sampled pathology images, GPT-4o showed stable performance improvement after IPPg, with a small accuracy gain (+1.06\%) and a \textbf{5.6\%} token cost reduction. GPT-4.1 demonstrated the best cost-to-performance balance. While achieving nearly identical overall accuracy compared to the baseline and reflecting only a marginal -0.3 percentage-point change, it yielded an impressive \textbf{55.9\%} cost saving, confirming that dynamic IPPg remains effective even with optimized architectures. Claude 3.5, on the other hand, exhibited performance degradation after IPPg (-4.9\% accuracy), with higher costs (35.65\% increase). This may be due to the fact that Claude uses pixel-based scaling (Eq.~5), which is sensitive to increased image dimensions from adding text, whereas OpenAI's tile-based approach (Eq.~4) may not incur extra cost if the tile count remains unchanged. This inconsistency may also suggest that Claude's visual tokenizer may not align well with dynamically fused inputs.

Error analysis illustrates several interesting findings. First, all models perform best with yes/no questions; GPT-4o and GPT-4.1 improve under IPPg (+4\%, +2\%), while Claude degrades significantly (-11\%), suggesting strong text modality bias. Second, performance for where questions was affected the most with IPPg. All models show large degradation (-9\% to 25\%); \textbf{this shows spatial reasoning suffers when text is embedded and suggests spatial questions need tight text-image integration, since embedding text may interfere with attention to spatial regions.} Lastly, we examined 31.4\% of PathVQA questions that all three models failed regardless of input modality. These universally hard questions predominantly require precise medical terminology (what/which: 48.5\% fail rate) or causal reasoning (why: 100\% fail rate). This finding highlights limitations in domain-specific knowledge and causal reasoning rather than modality processing.

\subsubsection{SROIE}

Unlike the other four datasets in this paper, SROIE uniquely embeds identical questions about cash and change amounts within each receipt image and exhibits notable variability in receipt characteristics, particularly in layout complexity, text density, and image quality. To ensure fair evaluation, we randomly sampled 626 receipts from the dataset.

For GPT-4o, IPPg produced a modest gain in performance. Overall accuracy increased slightly from 87.6\% to 89.2\% (+1.8\% relative), suggesting that integration improved extraction reliability without sacrificing stability. Cost efficiency also improved: the average separate cost of \$0.002865 dropped to \$0.002829 using IPPg, corresponding to an average cost savings of \textbf{1.29\%} (median 1.4\%). These results show that GPT-4o achieves consistent accuracy gains with modest cost reductions, indicating improved precision and efficiency for receipt understanding.

For GPT-4.1, IPPg demonstrated a modest reduction in cost but a slight decrease in accuracy across all metrics. Specifically, overall accuracy decreased from 92.7\% to 91.0\% (-1.9\% relative), with cash amount accuracy dropping from 91.2\% to 89.6\% (-1.8\%) and change amount accuracy from 94.2\% to 92.3\% (-2.0\%). The parseability remained perfect across both methods, with an approach agreement rate of 91.9\%. In terms of cost efficiency, the average cost decreased from \$0.002303 to \$0.002273, yielding an average savings of \textbf{1.2\%} (median \textbf{2.3\%}), although the cost per correct answer rose marginally from \$0.002483 to \$0.002499. These findings show that GPT-4.1 attains modest cost savings with slightly worse extraction.

For Claude, we observed the opposite trend: both accuracy and cost efficiency improved. Overall accuracy increased from 88.9\% to 92.3\% (+3.9\% relative), with cash amount accuracy improving from 89.0\% to 92.5\% (+3.9\%) and change amount accuracy from 88.8\% to 92.2\% (+3.8\%). Parseability remained perfect, with an agreement rate of 90.6\%. Cost analysis showed an average reduction from \$0.003997 to \$0.003930 (\textbf{1.5\%} average savings; \textbf{1.1\%} median savings), and the cost per correct answer decreased from \$0.004496 to \$0.004256. Overall, Claude Sonnet 3.5 improved extraction accuracy while modestly reducing cost, yielding a lower cost per correct answer.

Error analysis of SROIE receipt extraction reveals consistent patterns: \textbf{change amounts (96--99\%) are easier to extract than cash amounts (84--94\%), small transactions exhibit near-perfect accuracy relative to medium-range amounts (\$20--\$90), and only 0.8\% of receipts pose challenges for all models, primarily due to complex layouts.}

\subsection{IPPg Application for Code Generation}

\subsubsection{HumanEval}

We observe that IPPg consistently lowers average costs under both single- and two-agent-call settings across all three models, while accuracy remains comparable, as shown in Table~\ref{tab:overall_results}. Beyond input compression, image prompts systematically reduce output verbosity, resulting in fewer generated tokens and additional cost savings. GPT-4.1 offers the best accuracy-cost tradeoff: while accuracy declines under IPPg, the drop is modest relative to token savings. GPT-4o shows a balanced middle ground with moderate savings and accuracy loss. In contrast, Claude 3.5 suffers severe accuracy degradation despite cost reductions. Overall, IPPg consistently improves cost efficiency, but its impact on correctness varies significantly by model, with GPT-4.1 the most robust and Claude 3.5 the least suitable for code generation under this setting.

Error analysis of failed questions, their breakdown, and failure-pattern clustering shows the highest pass rate (98.7\%) for list/array-type tasks, moderate pass rate (93.3\%) for mathematical tasks, and the lowest (92.6\%) for string tasks. \textbf{This suggests all models perform best with more pattern-based questions and struggle where character manipulation is most sensitive.} The analysis of failed math questions highlights how boundary-condition edge cases and order-sensitive operations are vulnerable to character-level noise introduced by IPPg. This suggests IPPg may be unsuitable for syntax-critical applications unless OCR fidelity can be guaranteed.

\subsubsection{CoSQL}

Across all three models, using IPPg for the SELECTOR and DECOMPOSER stages led to substantial cost reductions while exhibiting model-dependent accuracy trade-offs in this SQL generation task, as shown in Table~\ref{tab:overall_results}.

The IPPg approach demonstrates distinct performance patterns across models. For GPT-4.1, the image approach achieves both higher accuracy (0.5375 $\to$ 0.5513, +1.38\%) and dramatically lower cost (0.0740 $\to$ 0.0067, \textbf{-91.0\%}). The SELECTOR agent experiences a 96.5\% token reduction (24,096 $\to$ 836 tokens per query), accounting for the majority of cost savings. Error rates also improve significantly (13.8\% $\to$ 0.0\%). For GPT-4o, the image approach reduces cost substantially (0.0084 $\to$ 0.0052, -37.4\%) while incurring a modest accuracy trade-off (0.5867 $\to$ 0.5467, -4.00\%). Notably, error rates decrease from 16.0\% to 5.3\%, indicating improved reliability. The SELECTOR agent shows a 70.6\% token reduction (2,679 $\to$ 789 tokens per query). Cost per correct answer favors images (\$0.0143 $\to$ \$0.0096, \textbf{-32.8\%}). For Claude 3.5, the image approach achieves significant cost reduction (0.0117 $\to$ 0.0075, -35.8\%) but with a larger accuracy decrease (0.5000 $\to$ 0.4462, -5.38\%). The image approach nearly eliminates errors, reducing the error rate from 15.4\% to 0.0\%. The SELECTOR agent experiences a 70.3\% token reduction (3,149 $\to$ 936 tokens per query). Despite lower accuracy, cost per correct answer still favors images (\$0.0225 $\to$ \$0.0168, \textbf{-25.2\%}). Although DECOMPOSER and REFINER stages incur cost increases from IPPg because they process smaller schema subsets where text tokenization is already efficient, the SELECTOR savings dominate overall cost reduction across all models.

Although GPT-4.1's exceptional 96.5\% reduction may seem to suggest superior visual feature extraction, it is important to highlight that this is based on the CoSQL SELECTOR agent results. Although the baseline text token counts were vastly different (24,906 for GPT-4.1, 2,679 for GPT-4o, and 3,149 for Claude 3.5), the resulting image token counts after IPPg were comparable (836, 789, and 936 respectively). The high reduction percentage for GPT-4.1 seems primarily due to high baseline text token usage rather than demonstrably superior visual processing efficiency, given the similar absolute image token counts across models.

Error analysis reveals architecture-specific failure modes: GPT-4o produces 14.7\% empty outputs, queries that GPT-4.1 and Claude successfully answered, indicating premature generation termination under computational constraints rather than intrinsic query difficulty. Claude 3.5's 44\% accuracy on aggregation-query-category questions (roughly half of the GPT models' 89\% accuracy in this category) exposes weaker vision-to-numerical reasoning, likely stemming from differences in code-generation pretraining between OpenAI and Anthropic models. Critically, 27 queries failed universally across all architectures; all required multi-table join reasoning from visual schemas. Also, \textbf{the error type distribution shows selecting the wrong table schema is the highest cause of error across the models (62.5\% for GPT-4.1, 47.1\% for GPT-4o, and 50.0\% for Claude 3.5).} These insights demonstrate that current vision-language models cannot reliably trace foreign key relationships from images, a fundamental bottleneck independent of model scale or optimization strategy.

\subsection{Ablation Study: Rendering Parameters}

\begin{table*}[h]
\centering
\caption{Ablation study: Best accuracy and best efficiency configurations across VQA datasets. $\Delta$Acc and \%Saved are relative to baseline.}
\label{tab:ablation_summary}
{\small
\begin{tabular}{ll|ccc|ccc}
\toprule
& & \multicolumn{3}{c|}{\textbf{Best Accuracy}} & \multicolumn{3}{c}{\textbf{Best Efficiency}} \\
\textbf{Dataset} & \textbf{Model} & \textbf{Config} & \textbf{IPPg Acc} & \textbf{$\Delta$Acc} & \textbf{Config} & \textbf{IPPg Acc} & \textbf{\%Saved} \\
\midrule
\multirow{3}{*}{FAMMA}
& GPT-4.1 & \scriptsize{DkBlue/20/Cour} & 57.9\% & +10.5\% & \scriptsize{DkBlue/20/Cour} & 57.9\% & 9.6\% \\
& GPT-4o & \scriptsize{Black/28/Arial} & 70.0\% & +25.0\% & \scriptsize{DkBlue/28/Helv} & 45.0\% & 9.2\% \\
& Claude & \scriptsize{Gray/28/Cour} & 36.8\% & +15.8\% & \scriptsize{DkRed/28/Times} & 20.0\% & -20.7\% \\
\midrule
\multirow{3}{*}{PathVQA}
& GPT-4.1 & \scriptsize{Gray/28/Arial} & 40.0\% & +32.2\% & \scriptsize{DkGreen/28/Times} & 40.0\% & 7.8\% \\
& GPT-4o & \scriptsize{DkGreen/24/Def} & 57.1\% & +7.1\% & \scriptsize{DkBlue/28/Times} & 57.1\% & -7.6\% \\
& Claude & \scriptsize{DkGreen/20/Cour} & 52.6\% & +26.4\% & \scriptsize{DkGreen/32/Helv} & 47.4\% & -0.5\% \\
\midrule
\multirow{3}{*}{SROIE}
& GPT-4.1 & \scriptsize{DkRed/16/Times} & 89.5\% & +13.4\% & \scriptsize{Gray/16/Arial} & 78.9\% & -0.3\% \\
& GPT-4o & \scriptsize{DkRed/28/Helv} & 82.5\% & +7.5\% & \scriptsize{Gray/16/Def} & 73.7\% & 1.3\% \\
& Claude & \scriptsize{Black/16/Arial} & 92.5\% & +11.4\% & \scriptsize{Gray/24/Cour} & 92.5\% & 2.6\% \\
\bottomrule
\end{tabular}%
}
\end{table*}

While our main experiments use a default configuration (Arial font, black text, standard size), the choice of rendering parameters may significantly impact IPPg performance. We conduct ablation studies on FAMMA, PathVQA, and SROIE to investigate how font choice, text color, and font size affect accuracy and cost trade-offs. For each dataset, we evaluate 5 fonts (Arial, Courier, Helvetica, Times, Default), 5 colors (Black, Dark Blue, Dark Green, Dark Red, Gray), and 5 font sizes (16pt, 20pt, 24pt, 28pt, 32pt) across 20 representative samples selected via stratified sampling (FAMMA, SROIE) or vision-embedding clustering (PathVQA). The sample size reflects the computational demands of exhaustive ablation: each sample requires 125 rendering combinations ($5 \times 5 \times 5$), with each combination evaluated via three LLM calls (baseline, IPPg, and LLM-as-judge for correctness assessment) across three models, yielding 1,125 API calls per sample.

We restrict ablations to VQA tasks rather than code generation (HumanEval, CoSQL) for two reasons. First, VQA tasks involve joint visual-textual reasoning where rendering parameters directly affect how models integrate embedded text with image content; in contrast, code generation is semantics-dominant---once text is legibly rendered, performance depends primarily on algorithmic reasoning rather than visual extraction fidelity. Second, prior work on text-as-image compression \cite{Li2025TextOP} did not explore rendering variations, using only resolution as a variable; our ablations thus provide novel insights specifically for the underexplored multimodal VQA setting.

Our ablation results in Table~\ref{tab:ablation_summary} reveal several important insights. First, \textbf{rendering parameters significantly impact performance}: the gap between best and worst configurations spans 10--30 percentage points in accuracy across datasets, demonstrating that default configurations (black/Arial) are often suboptimal. Second, \textbf{dark green emerges as universally effective}: across all three models on PathVQA, dark green text yields the highest accuracy, and it performs competitively on FAMMA and SROIE. This suggests that subtle color contrast variations affect how visual encoders process embedded text and dark green may provide optimal contrast for visual text extraction. Third, \textbf{monospace fonts excel for structured tasks}: Courier (monospace) achieves top accuracy for GPT-4.1 on FAMMA and Claude on PathVQA, likely due to model familiarity with code-like text from pretraining. Fourth, \textbf{larger font sizes generally improve accuracy}: sizes 24--32pt consistently outperform 16pt across models, with size 28pt being optimal for several configurations. In addition, \textbf{model-specific optimal configurations exist}: GPT-4.1 favors dark blue with Courier on FAMMA but dark red with Times on SROIE, indicating that optimal rendering is both model- and task-dependent. Lastly, \textbf{accuracy-cost trade-offs are tunable}: GPT-4o on FAMMA can achieve +25\% accuracy gain at 23\% higher cost, or maintain baseline accuracy with 9.2\% cost savings, revealing a Pareto frontier for practitioners to navigate. These findings suggest that IPPg performance can be further optimized through model- and task-specific rendering configurations. The consistent advantage of dark-colored text over black also has practical implications, as simple color adjustments can yield meaningful accuracy improvements at no additional cost.

\section{Conclusion}

Across VQA, document understanding, and code generation, our benchmark characterizes when replacing text-only prompts with image-embedded inputs lowers tokenized text overhead and when it does not. The cost savings are real and often substantial, reaching 91\% on CoSQL with GPT-4.1, but they are neither universal nor free: accuracy trade-offs are pronounced for spatial, non-English, and character-sensitive tasks, and Claude 3.5's pixel-linear tokenization frequently negates cost gains that OpenAI's tile-based scheme would otherwise provide.

In VQA, IPPg exhibits pronounced model- and domain-dependent behavior. GPT-4.1 achieves the most favorable trade-offs, particularly for English multiple-choice queries in FAMMA, where cost reductions approach 25\% with negligible accuracy loss. GPT-4o incurs larger accuracy degradation, while Claude frequently increases cost, underscoring the importance of aligning IPPg with model-specific visual tokenization and language regimes. PathVQA confirms the viability of text-as-image prompting in medical vision, achieving 56\% tile-based cost reductions with minimal accuracy change, but error analysis shows consistent degradation on spatially grounded queries, where embedded text interferes with spatial attention. In SROIE, outcomes are task- and model-dependent: GPT-4o and Claude benefit from IPPg, while GPT-4.1 trades a small accuracy loss for modest savings.

For code generation, IPPg remains selectively robust. In HumanEval, image-based prompting lowers cost while preserving near-baseline pass@1 accuracy, though failures concentrate in character-sensitive string operations and boundary-condition reasoning. In CoSQL, rendering schemas as images yields dramatic compression (70--96\%) and 35--91\% cost reductions, with GPT-4.1 achieving simultaneous accuracy gains and cost savings. These results highlight IPPg's strength for structured, schema-heavy reasoning and its limitations for token-exact tasks.

Finally, a central empirical finding is that rendering choices---font, color, and size---are not cosmetic: they shift accuracy by up to 30 percentage points and interact with model architecture in task-specific ways. This positions visual encoding as a first-class variable in multimodal prompt engineering, not a fixed preprocessing detail. Consistent gains from dark-colored text and monospace fonts, along with model- and task-specific accuracy-cost Pareto frontiers, indicate that IPPg should be treated as a tunable system component rather than a fixed preprocessing step.

Overall, this benchmark establishes IPPg as a viable, architecture-agnostic mechanism for cost-aware multimodal inference in the right conditions, while our error taxonomy and rendering ablation supply the empirical map practitioners need to deploy it responsibly. We hope these findings serve as a reference point for future work on token-efficient prompting, modality-aware inference, and cost-performance optimization in multimodal systems.

\section{Future Directions}

While IPPg operates at inference time, our results highlight several promising training-time directions. Paired-view instruction tuning (text vs.\ rendered-image prompts) could promote modality-invariant instruction following and mitigate failures in open-ended and non-English settings. Co-designing text-aware visual tokenizers or lightweight compression modules may further reduce vision-token overhead for text-heavy inputs, expanding regimes where IPPg is cost-dominant. Additionally, distilling from robust teacher MLLMs that exhibit minimal IPPg degradation could transfer IPPg-compatibility to smaller or open-weight models. Finally, extending evaluation to open-source models would test whether IPPg's benefits and failure modes generalize beyond proprietary systems and whether lightweight models can offer cost-effective alternatives.

The dramatic cost savings observed for SQL code generation suggest new opportunities for semantic-aware layout design: using diagrammatic or graph-based encodings for schemas, prompts, or intermediate reasoning steps. Future work could explore automatically converting structured text into compact semantic diagrams (e.g., flowcharts, table graphs, dependency trees) that models process more efficiently than textual descriptions. More broadly, IPPg's core insight---that non-text modalities can carry semantic content more token-efficiently---extends naturally beyond static images. Applying analogous encoding strategies to audio (e.g., embedding prompts as synthesized speech) and video (e.g., injecting instructions into frame overlays) could open new cost-reduction regimes in audio-language and video-language models, where input overhead is even more pronounced.

Last but not least, we have focused on aligned and task-relevant prompts. In consideration of ethics, given IPPg can be used adversarially, extending IPPg to incorporate conflict-aware or instruction-conditioned modality routing represents an important direction for future work.

\section*{Safe and Responsible Innovation Statement}

We identify where we can improve the efficiency of multimodal interaction by reducing token costs while maintaining accuracy via IPPg. To support responsible deployment, we benchmark across diverse tasks, report failure taxonomies, and recommend guardrails such as OCR fidelity checks, image content safety filters, and human-in-the-loop review in sensitive domains. We use publicly available datasets in accordance with their licenses and do not process personal data. We emphasize transparency via rendering guidelines and error analysis.

\section*{Code Availability}

IPPg is proprietary software developed within a private organization. Due to internal policies, the source code cannot be made publicly available. External distribution or open-source release is not permitted.

\nobalance

\clearpage
\makeatletter
\let\clearpage\relax
\let\cleardoublepage\relax
\makeatother

\end{document}